\documentclass{article}
\usepackage{amsmath,graphicx,mlspconf}
\usepackage{amssymb}
\newcommand{\eg}{\emph{e.g.}}
\newcommand{\ie}{\emph{i.e.}}
\newcommand{\etal}{\emph{et~al.}}
\DeclareMathOperator{\E}{\mathbb{E}}

\toappear{Accepted to 2017 IEEE International Workshop on Machine Learning for Signal Processing}
\title{Adversarial nets with perceptual losses for text-to-image synthesis}
\name{Miriam Cha, Youngjune Gwon, H.~T.~Kung}
\address{Harvard University}
\begin{document}
%\ninept
%

\maketitle
\begin{abstract}
Recent approaches in generative adversarial networks (GANs) can automatically synthesize realistic images from descriptive text. Despite the overall fair quality, the generated images often expose visible flaws that lack structural definition for an object of interest. In this paper, we aim to extend state of the art for GAN-based text-to-image synthesis by improving perceptual quality of generated images. Differentiated from previous work, our synthetic image generator optimizes on perceptual loss functions that measure pixel, feature activation, and texture differences against a natural image. We present visually more compelling synthetic images of birds and flowers generated from text descriptions in comparison to some of the most prominent existing work. 
\end{abstract}

\section{Introduction}
The recently introduced generative adversarial net (GAN) \cite{goodfellow2014} is designed to benefit from the competition between a pair of simultaneously trained learning models with opposite goals. In the training of a generative model, GAN employs a discriminative model to provide a feedback (\ie, the discriminator's prediction output) crucial in computing the generator's loss function. Desirably, when the training reaches to a steady-state equilibrium where the discriminator confusion is at maximum on synthetic data from the generator, one can conclude the overall GAN objective accomplished. 

In this paper, we aim to extend the GAN framework for automatic text-to-image synthesis. We are interested in generating a perceptually high-quality image that matches a descriptive text input. In particular, we address two important problems related to cross-modal translation and realistic image production. Good cross-modal translation results in a generated image closely matching the given text description whereas a realistic image can hardly be distinguished from natural images. 

To ensure good cross-modal translation, we adopt a \emph{contextual} loss term in the generator following the conditional GAN framework \cite{mirza2014conditional}. To generate realistic images, we in addition introduce \emph{perceptual} loss terms for the generator, corresponding to pixel, feature activation, and texture reconstruction losses. Thus our approach is to regularize the original minimax optimization for GAN with both contextual and perceptual loss terms. 

\begin{figure}[t]
\centering
\includegraphics[width=.3\textwidth]{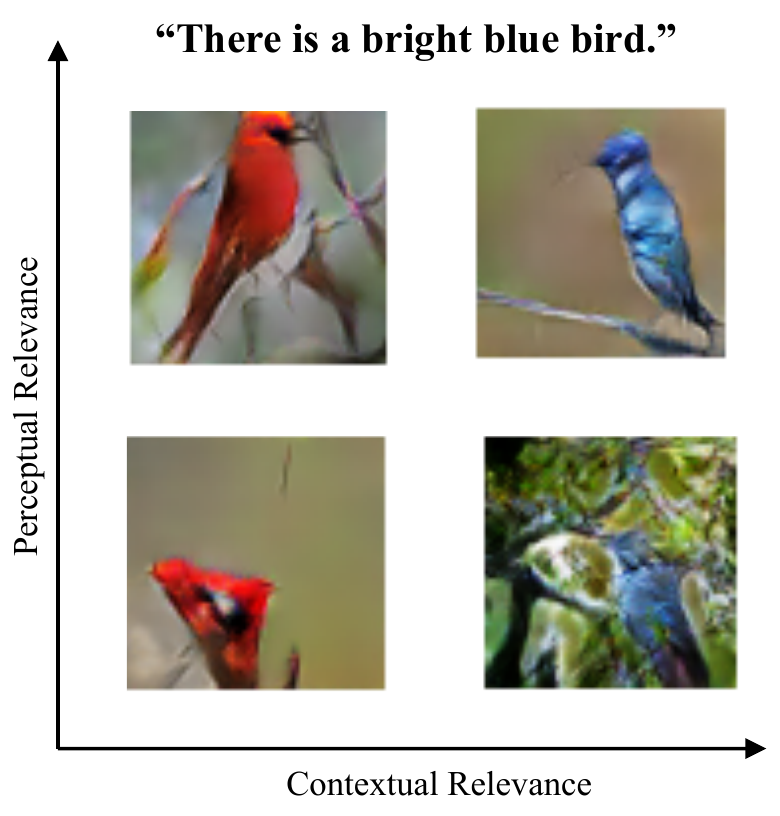}
\caption{Four synthetic bird images generated for the given text, displaying different levels of contextual and perceptual relevance. The upper right image is most desirable, as it has higher relevance on both accounts.}
\label{fig:contperc}
\end{figure}

We illustrate the effect of contextual and perceptual losses in Figure~\ref{fig:contperc}. Here, the task is to generate an image for text, ``There is a bright blue bird." The two axes in the figure indicate the degree of contextual and perceptual relevances. The red birds on the left are contextual mismatches because the input text mentions only `blue' bird. Hinted by higher perceptual relevance, the bird images on the top have more natural and better overall perceptual quality than the bottom. We are interested in synthesizing images in the upper right corner which have higher relevance in both context and perception.

The use of GAN to generate realistic images has been attempted by others. Radford~\etal~\cite{radford2015unsupervised} propose deep convolutional generative adversarial net (DCGAN) that takes in a random noise vector to generate synthetic images. We base our approach on conditional GAN by Mirza \& Osindero \cite{mirza2014conditional}. By conditioning both generator and discriminator networks on the text input, conditional DCGAN can model the phenomenon that many different natural images can be mapped to the same text description. 

Several encouraging results follow in recent literature. Gauthier~\cite{gauthier2014conditional} applies conditional GAN to generating MNIST and facial images by using class labels as side information. Denton~\etal~\cite{denton2015deep} introduce a Laplacian pyramid extension to conditional GAN that can generate an image at different resolutions. The notion of perceptual loss in image processing is originated by Gatys \etal~\cite{gatys2015texture}. They have proposed to use the Gram matrix of convolutional neural net activations (\ie, high-level image feature vectors) between two images to make artistic style transfer possible. In Johnson \etal~\cite{johnson2016perceptual}, the same perceptual loss function is used for image super-resolution. Ledig \etal~\cite{ledig2016photo} propose super-resolution GAN (SRGAN) that combines per-pixel, VGG, and adversarial losses as the perceptual loss function. For VGG loss, they take the ReLU activation layers of the pre-trained 19-layer convolutional neural net by Oxford's Visual Geometry Group (VGG) \cite{simonyan2014very}.

Reed \etal~\cite{reed2016generative} present an architecture based on conditional DCGAN. They have trained bird and flower image generators conditioned on text features computed from a character-level recurrent neural net. Our work builds on Reed \etal~\cite{reed2016generative}. We tackle the same bimodal text-to-image task and use the same datasets. However, our generator minimizes on a perceptual loss term given by the Gram, per-pixel, or VGG loss function alongside the contextual loss obtained from the discriminator in conditional GAN. Ledig \etal~have explored the per-pixel, VGG, and adversarial loss functions on the image super-resolution task, not text-to-image synthesis.

The rest of this paper is organized as follows. In Section 2, we provide a background on generative adversarial net, covering its variants and mathematical formulations. In Section 3, we describe our conditional GAN approach incorporating the perceptual loss in addition to the contextual loss. Section 4 presents an experimental evaluation of our approach trained on the Caltech-UCSD Bird and Oxford-102 flower datasets. Section 5 concludes the paper.

\section{Background}
In this section, we review adversarial training techniques for a generative model with an emphasis on its conditional variant, which can be thought as a multimodal extension.

\subsection{Generative adversarial nets (GAN)}
Goodfellow \etal~\cite{goodfellow2014} introduce GAN as a new means to learn probability distributions for data. GAN constitutes a generative model $G$ and a discriminative model $D$ that are simultaneously trained in the competition described by a two-player minimax game:
\begin{equation} \label{eq:gan} \small
\min_G\max_D \E_{\mathbf{x} \sim p_{\text{data}}}[\log D(\mathbf{x})] + \E_{\mathbf{z} \sim p_\mathbf{z}}[\log(1-D(G(\mathbf{z})))] 
\end{equation}
Here, the objective of $G$ is to produce a data estimate $\hat{\mathbf{x}}$ to the real $\mathbf{x}$, using a latent input variable $\mathbf{z}$ (\eg, noise) with a prior distribution $p_{\mathbf{z}}$. On the other hand, $D(\mathbf{x})$ represents the probability that $\mathbf{x}$ originates from the real data distribution $p_{\text{data}}$. Hence, $D(G(\mathbf{z}))$ can be used to evaluate the quality of generated data $\hat{\mathbf{x}}=G(\mathbf{z})$ with respect to the real $\mathbf{x}$. In practice, $G$ and $D$ are typically implemented as neural nets. We can train $G$ and $D$ by backpropagation.

\subsection{Conditional generative adversarial nets (CGAN)}
Mirza \& Osindero \cite{mirza2014conditional} extend GAN by conditioning $G$ and $D$ on side information. For CGAN, Eq. (\ref{eq:gan}) is rewritten as 
\begin{equation} \label{eq:cgan} \small
\min_G\max_D \E_{\mathbf{x} \sim p_{\text{data}}}[\log D(\mathbf{x}|\mathbf{y})] + \E_{\mathbf{z} \sim p_\mathbf{z}}[\log(1-D(G(\mathbf{z}|\mathbf{y})))]
\end{equation} where the extra information $\mathbf{y}$ can be a class label or data from another modality. For the latter case, since certain modalities are often observed together (\eg, audio-video and image-text), it is convenient to use CGAN. In neural net implementation, $G$ takes in $\mathbf{z}|\mathbf{y}$ as a joint representation that concatenates $\mathbf{z}$ and $\mathbf{y}$ into a single input vector. Similarly, another joint representation is used to train $D$. 

\subsection{Deep convolutional neural networks for GAN}
Convolutional neural nets (CNNs) remain to be state-of-the-art for visual recognition tasks. Radford \etal~\cite{radford2015unsupervised} present convincing evidence in favor of deep CNN as a strong candidate architecture for adversarial learning. Their DCGAN has shown to learn a hierarchy of representations from object parts to scenes for image generation tasks. According to Radford \etal, the success of DCGAN is attributed to their three architectural modifications to CNN. First, strided convolutions replace deterministic spatial pooling (\eg, max and average pooling) to learn spatial down- and upsampling. Secondly, fully connected layers following convolutional layers are removed to allow deeper representations. Lastly, batch normalization, which conditions the input to have zero mean and unit variance, induces to learn more useful features. In this work, we adopt the \texttt{DCGAN.torch} framework\footnote{https://github.com/soumith/dcgan.torch} to implement convolutional generator and discriminator.

\begin{figure*}[t]
\centering
\includegraphics[width=.89\textwidth]{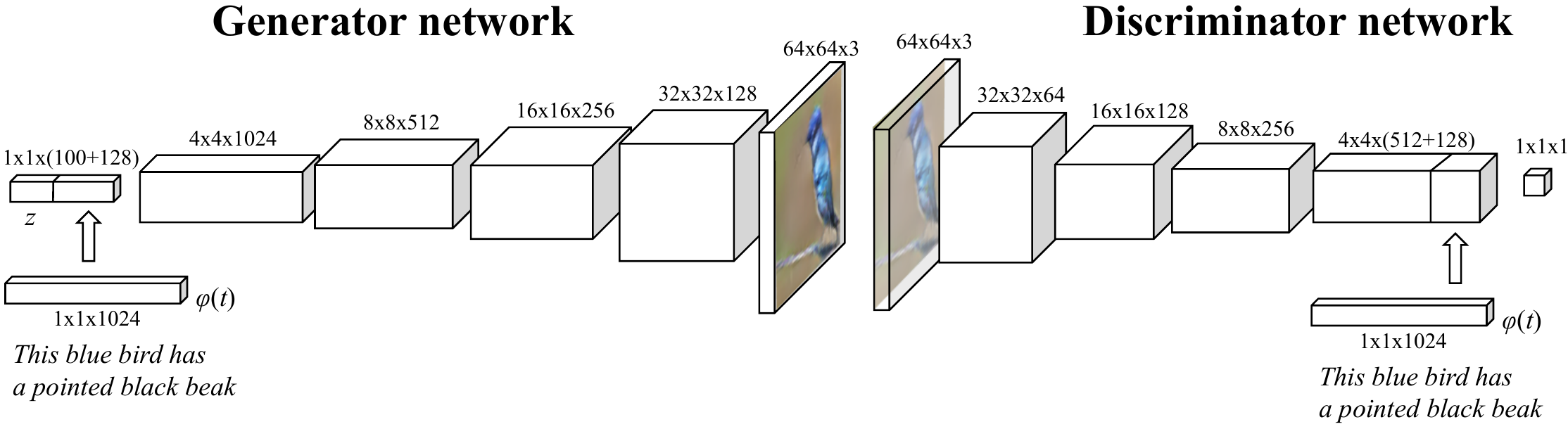}
\caption{Architecture of the generator and discriminator networks of our conditional GAN synthesis model.}
\label{fig:ours}
\end{figure*} 

\section{Approach}
Our task is to generate realistic images that match text input. At a high-level, we build a DCGAN as in Reed \etal~\cite{reed2016generative} and train it with contextual and perceptual loss terms by conditioning on the input text. The main contribution of our approach is the inclusion of additional perceptual loss in training the generator. In particular, we compare the use of three different perceptual loss functions based on pixel, activation, and texture reconstruction. This section first presents our network architecture and then describes the formulation of loss functions used to train our DCGAN.

\subsection{Architecture}
Figure~\ref{fig:ours} depicts our network architecture. We use input text encoding $\varphi(t)$ at both generator and discriminator networks. In the generator network $G$, $\varphi(t)$ (in reduced dimensionality) is concatenated with a noise sample $\mathbf{z}$ and propagated through stages of fractional-strided convolution processing. In the discriminator network $D$, an input image is processed through layers of strided convolution before concatenated to $\varphi(t)$ (in reduced dimensionality) for computing the final discriminator score. Both $G$ and $D$ are optimized by gradient descent on their loss functions $\mathcal{L}_G$ and $\mathcal{L}_D$, respectively. 

We denote $\mathbf{x}$ for image data, $t$ for text data, $\mathbf{h} = \varphi(t)$ for text encoding, and $\mathbf{z}$ for noise input. A superscript $\!^{(i)}$ is used to designate $i$th example. For our task, a training example is an image-text pair. When applying $m$ training examples (\eg, mini-batches) $\{(\mathbf{x}^{(1)},\mathbf{h}^{(1)}),...,(\mathbf{x}^{(m)},\mathbf{h}^{(m)})\}$ with noise input $\{\mathbf{z}^{(1)}, ..., \mathbf{z}^{(m)}\}$ sampled from the prior $p_z$, $G$ generates corresponding synthetic images $\{\hat{\mathbf{x}}^{(1)}, ..., \hat{\mathbf{x}}^{(m)}\}$. 

%\subsection{Discriminator loss}
The discriminator loss function is given by
{\small \begin{align}
\label{eq:Ld_loss}
\mathcal{L}_D = &- \frac{1}{m} \sum_i \log(D(\mathbf{x}^{(i)}|\mathbf{h}^{(i)})) \nonumber\\ 
& - \frac{1}{2}[\log (1-D(\mathbf{x}^{(i)}|\hat{\mathbf{h}}^{(i)})) + \log (1-D(\hat{\mathbf{x}}^{(i)}|\mathbf{h}^{(i)}))]
\end{align}}where $\{\mathbf{x}^{(i)}|\mathbf{h}^{(i)}\}$ corresponds a real image with its corresponding right text, $\{\mathbf{x}^{(i)}|\hat{\mathbf{h}}^{(i)}\}$ real image with arbitrary text (randomly chosen from a different image category), and $\{\hat{\mathbf{x}}^{(i)}|\mathbf{h}^{(i)}\}$ synthetic image with right text. 

%\subsection{Generator loss}
The generator loss function is a weighted sum of two parts
\begin{equation}
\label{eq:Lg_loss}
\mathcal{L}_G = \ell_\text{cont} + \lambda \ell_\text{perc}
\end{equation} where $\ell_\text{cont}$ is contextual loss, $\ell_\text{perc}$ perceptual loss, and a weight parameter $\lambda$.

\subsection{Contextual loss}
The contextual loss used in our implementation is:
\begin{equation}
\label{eq:L_cont}
\ell_\text{cont} = - \frac{1}{m} \sum_i \log(D(G(\mathbf{z}^{(i)} | \mathbf{h}^{(i)})))
\end{equation}
where $D(G(\mathbf{z} | \mathbf{h}))$ is the probability that the generated image and the conditioned text form a real contextual pair. For better gradient behavior, we minimize $-\log(D(G(\mathbf{z} | \mathbf{h})))$ rather than minimizing the original $\log (1-D(G(\mathbf{z} | \mathbf{h})))$ \cite{goodfellow2014}.

\subsection{Perceptual loss}
The contextual (adversarial) loss for the generator, as presented in Reed \etal~\cite{reed2016generative}, only considers the semantic relatedness between images and text descriptions without an explicit regularization term that penalizes large visual changes between the synthetic and real images. In contrast, in addition to the contextual loss term, we adopt a perceptual loss $\ell_\text{perc}$ as a part of the loss in training the generator. We propose three perceptual loss functions each aiming to enforce perceptual similarity between the real and the generated images. That is, by minimizing the perceptual loss the visual difference between the synthetic and the real images can be lowered. This helps preserve the perceptual realism. 
\\
\textbf{Pixel reconstruction loss.} Making image adjustment to minimize pixel-wise losses is a simple approach to encourage visual similarity between images. The pixel reconstruction loss calculates the mean squared error between a real image $\mathbf{x}^{(i)}$ and a corresponding synthetic image $\hat{\mathbf{x}}^{(i)}$ as
\begin{equation}
\label{eq:pixelloss}
\ell_\text{perc\_Pix} = \frac{1}{m} \sum_i \Arrowvert \mathbf{x}^{(i)} - \hat{\mathbf{x}}^{(i)}\Arrowvert_{2}^{2} 
\end{equation} 
where $\Arrowvert \cdot \Arrowvert_2$ is the $\ell_2$ norm. The pixel reconstruction loss function encourages the pixels of the two images to match. Unlike image super-resolution, text-to-image synthesis involves one-to-many mapping between two different data kinds. Thus, we find usage of high-level image features more appropriate. \\
\textbf{Activation reconstruction loss.} Instead of promoting pixel-wise match between synthetic and real images, we can encourage high-level feature representations of the images to be similar. Let $\mathbf{A}^j_{\mathbf{x}^{(i)}}$ or $\mathbf{A}^j_{\hat{\mathbf{x}}^{(i)}}$ be the rectified linear unit (ReLU) activation outputs of the $j$th convolutional layer within a classification network such as VGG when $\mathbf{x}^{(i)}$ or $\hat{\mathbf{x}}^{(i)}$ is used as an input, respectively. The feature reconstruction loss is defined as
\begin{equation}
\label{eq:vggloss}
\ell_\text{perc\_VGG} = \frac{1}{m} \sum_i \Arrowvert \mathbf{A}^j_{{\mathbf{x}^{(i)}}} - \mathbf{A}^j_{\hat{\mathbf{x}}^{(i)}} \Arrowvert_{F}^{2} 
\end{equation} 
where $ \Arrowvert \cdot  \Arrowvert_F$ is the Frobenius norm. Here, we use a 19-layer VGG network pretrained on the ImageNet dataset \cite{russakovsky2015imagenet}. Activation outputs derived from high levels capture image content and overall structures such as object shapes that may be useful for classifying objects. By minimizing the differences in the activation outputs, we encourage the generated image to be classified similarly as the real image, thereby containing objects of the same class as those in the real image. 
 \\ 
\textbf{Texture reconstruction loss.}
Although image content and overall structures are well captured in the activation outputs, style-related features such as texture and recurring patterns may not. In order to capture whether the generated image and the real image use combinations of nearly identical set of supporting filters, we compare Gram matrices of the activation outputs, as previously being proposed for image super-resolution \cite{johnson2016perceptual} and style transfer \cite{gatys2015texture,gatys2015neural}:
\begin{equation}
\label{eq:gramloss}
\ell_\text{perc\_Gram} = \frac{1}{m} \sum_i \Arrowvert S(\mathbf{A}^j_{{\mathbf{x}^{(i)}}}) - S(\mathbf{A}^j_{\hat{\mathbf{x}}^{(i)}}) \Arrowvert_{F}^{2} 
\end{equation} 
where $S(\mathbf{A}^j_{{\mathbf{x}^{(i)}}})$ and $S(\mathbf{A}^j_{{\hat{\mathbf{x}}^{(i)}}})$ are the Gram matrix of activation vectors resulting from filters at $j$th convolutional layer for input $\mathbf{x}^{(i)}$ and $\hat{\mathbf{x}}^{(i)}$, respectively. Note that since the Gram matrix captures the correlations between these filters, it effectively describes an object-level style of an image.

\section{Experiments}

\begin{figure*}[t]
\centering
\includegraphics[width=0.95\textwidth]{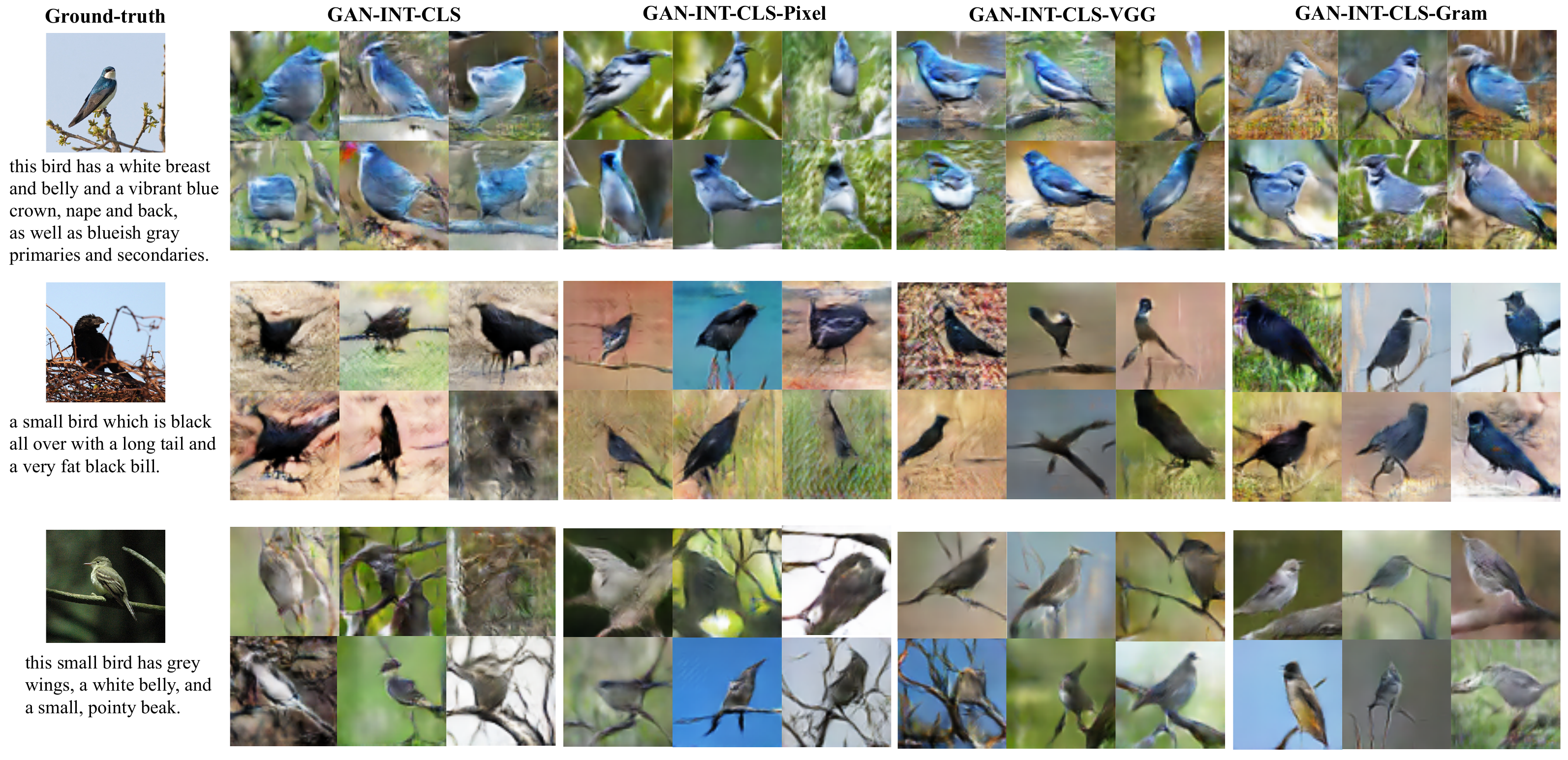}
\caption{Generated bird images by conditioning on text from unseen test categories using the baseline GAN-INT-CLS \cite{reed2016generative} and the baseline with three different perceptual loss functions: pixel (GAN-INT-CLS-Pixel), activation (GAN-INT-CLS-VGG), and texture (GAN-INT-CLS-Gram) reconstruction losses.}
\label{fig:comp}
\end{figure*}

\begin{figure*}[t]
\centering
\includegraphics[width=0.95\textwidth]{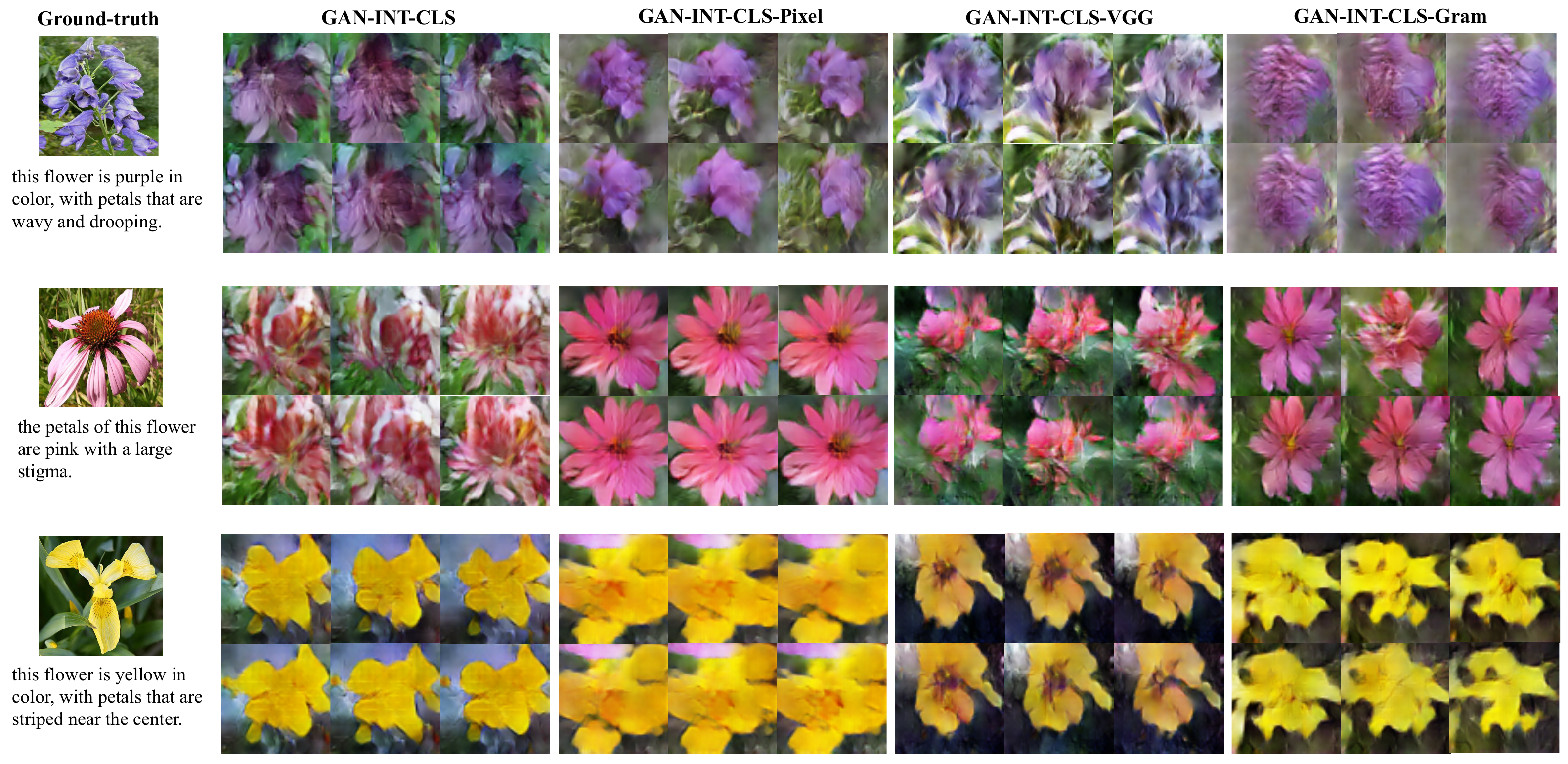}
\caption{Generated flower images by conditioning on text from unseen test categories using the baseline GAN-INT-CLS \cite{reed2016generative} and the baseline with three different perceptual loss functions: pixel (GAN-INT-CLS-Pixel), activation (GAN-INT-CLS-VGG), and texture (GAN-INT-CLS-Gram) reconstruction losses.}
\label{fig:comp2}
\end{figure*} 

We evaluate the proposed models with perceptual losses on Caltech-UCSD Bird (CUB) \cite{wah2011caltech} and Oxford-102 flower \cite{nilsback2008automated} datasets. CUB consists of 11,788 images of 200 bird species. The Oxford-102 has 8,189 images of flowers from 102 different types. A recent extension of these datasets has collected 10 visual description sentences for each image \cite{reed2016learning}. We follow train/test split by Reed~\etal~\cite{reed2016learning}. That is, the images in CUB are split into 150 training/validation and 50 test classes. The flower dataset is split into 82 training/validation and 20 test categories. Thus during testing, an input text is from a category not present in the training set.

All images are resized to 64$\times$64$\times$3. For word representation, we adopt the state-of-the-art pre-trained character-based embedding on the visual descriptions, namely character-level ConvNet with a recurrent neural network (char-CNN-RNN), which outputs text embedding vectors of dimensionality 1,024. As shown in Fig.~\ref{fig:ours}, both the generator and discriminator are deep convolutional neural networks. The text embedding is projected to a 128-dimensional vector by linear transformation. In the generator, the input is formed by concatenating this text embedding vector with a 100-dimensional noise sampled from unit normal distribution. In the discriminator, the projected text-embedding vector is depth concatenated with the final convolutional feature map. 

All models are trained with mini-batch stochastic gradient descent with a mini-batch size of 64. We adopt the ADAM optimizer \cite{kingma2014adam} with momentum 0.5 and learning rate 0.0002 as used in Radford~\etal~\cite{radford2015unsupervised}. We use for Eq.~(\ref{eq:Lg_loss}) the hyperparameter $\lambda=10^{-6}$ to balance contextual and perceptual losses.  

In the experiments, we evaluate our models based on human and machine judgments. For human judgment, we qualitatively evaluate the generated synthetic images given query text descriptions by visual inspection. For machine judgment, we input the generated image to an object classifier to assess quantitatively the realism of the synthetic images. 

\subsection{Qualitative evaluation}
Fig.~\ref{fig:comp} shows qualitative examples comparing our results with the results of baseline method (GAN-INT-CLS) by Reed~\etal~\cite{reed2016generative} for the CUB dataset. The generated bird images are conditioned on text from unseen test categories. Our three methods are the baseline combined with pixel (GAN-INT-CLS-Pixel), activation (GAN-INT-CLS-VGG), and texture (GAN-INT-CLS-Gram) reconstruction losses. In these results, it is clear that GAN-INT-CLS is aware of the semantic relatedness between images and text. However, the bird images by GAN-INT-CLS generally lack structural definition. The baseline method with pixel reconstruction loss (GAN-INT-CLS-Pixel) yields slightly better bird structures. We find that GAN-INT-CLS-VGG and  GAN-INT-CLS-Gram yield the perceptually and contextually most compelling results. We speculate that the improvements stem from the feature maps that focus on the object-specific structures while leaving the contextual loss focusing on semantic relatedness. 

In Fig.~\ref{fig:comp2}, we present results on the Oxford-102 flower dataset using the baseline and our methods. As shown, all four methods have generated plausible looking flower images according to the query text. The generated flowers look like flowers from different classes. This may be due to one-to-many problem where the same text can be used to describe many different classes. In addition, the descriptions provided by human may not have been descriptive enough to distinguish between different types of flowers.   

\subsection{Quantitative evaluation}
For quantitative evaluation, we utilize GoogLeNet \cite{szegedy2015going} to evaluate how well the generated images are correctly classified as birds or flowers. We use the pretrained GoogLeNet model from \texttt{inception-v3.torch}.\footnote{https://github.com/Moodstocks/inception-v3.torch} For the evaluation, we input three query texts: ``the bright blue bird has a white colored belly,'' ``this bird is yellowish orange with black wings,'' and ``this vibrant red bird has a pointed black beak.'' For each text query, we generate 500 synthetic bird images and input the total 1,500 such images to GoogLeNet to do bird classification. Since there are many different types of birds from 1,000 object classes of the GoogLeNet output, we use a world list of birds\footnote{http://ces.iisc.ernet.in/hpg/envis/sibleydoc63.html} to make a conglomerate bird classifier, such that all types of birds are classified as a bird. 

We perform similar tasks for the flower dataset and evaluate the performance of flower classification. We use query texts, ``this flower has white petals and a yellow stamen,'' ``the center is yellow surrounded by wavy dark purple petals,'' and ``this flower has lots of small round pink petals.'' A comprehensive list of flowers is obtained from Lyons \cite{lyons1900plant}.

Table~\ref{tab:quant_comp} presents the GoogLeNet bird and flower classification results on images generated by GAN-INT-CLS, GAN-INT-CLS-Pixel, GAN-INT-CLS-VGG, and GAN-INT-CLS-Gram. We observe the trend that more images are correctly classified as birds and flowers when perceptual loss functions are used. The quantitative evaluation confirms the qualitative improvements observed in Fig.~\ref{fig:comp} and Fig.~\ref{fig:comp2}, showing a combination of contextual and perceptual loss terms generates better images for both human perception and machine classification. We notice that for the flower case our results using GAN-INT-CLS-Gram are significantly better than other methods. This can be explained by observing that there are non-flower objects such as hen-of-the-woods and earthstar, which have subtle differences in styles as compared to flowers that humans can pick up. However, without incorporating a proper perceptual loss in the generator, GAN will not be able to synthesize flowers with distinctive features.  

\begin{table}[t]
\centering
\small
\caption{GoogLeNet bird and flower classification accuracies}
\label{tab:quant_comp}
\begin{tabular}{r||c|c}
\hline
\multicolumn{1}{l||}{}      & \textbf{Bird} & \textbf{Flower} \\ \hline
\textbf{GAN-INT-CLS}       & 0.76 & 0.66  \\ \hline
\textbf{GAN-INT-CLS-Pixel} & 0.83              &    0.66             \\ \hline
\textbf{GAN-INT-CLS-VGG}   & 0.80              &    0.76         \\ \hline
\textbf{GAN-INT-CLS-Gram}  & 0.85          &      0.89     \\ \hline
\end{tabular}
\end{table}

\section{Conclusion}
In this paper, we have described GAN-based text-to-image synthesis methods that use both contextual and perceptual losses. The contextual loss in existing GAN literature focuses on semantic relatedness between text and image, whereas the proposed perceptual loss focuses on the object-specific structure. Our results on CUB and Oxford-102 datasets suggest that adopting perceptual loss functions is helpful for improving visual realism in text to image synthesis. In future work, we hope to explore different conditional GAN frameworks and investigate additional perceptual loss functions.

\footnotesize
\section{Acknowledgements}
This work is supported by the MIT Lincoln Laboratory Lincoln Scholars Program and in part by gifts from the Intel Corporation and the Naval Supply Systems Command award under the Naval Postgraduate School Agreements No.
N00244-15-0050 and No. N00244-16-1-0018. 

% Below is an example of how to insert images. Delete the ``\vspace'' line,
% uncomment the preceding line ``\centerline...'' and replace ``imageX.ps''
% with a suitable PostScript file name.
% -------------------------------------------------------------------------

\vfill
%\pagebreak

%\clearpage
\footnotesize
%\tiny
\bibliographystyle{IEEEbib}
\bibliography{paper}

\end{document}